\documentclass[journal]{IEEEtran}

\usepackage{times}
\usepackage{soul}
\usepackage{url}
\usepackage[hidelinks]{hyperref}
\usepackage[utf8]{inputenc}
\usepackage[small]{caption}
\usepackage{graphicx}
\usepackage{amsmath}
\usepackage{booktabs}
\usepackage{multirow}
\usepackage{amsthm}
\usepackage{verbatim}
\usepackage{amsfonts,amssymb,mathrsfs}
\usepackage[subfigure]{tocloft}
\usepackage{subfigure}
\usepackage{xcolor}

\usepackage{algorithmic}
\usepackage[hidelinks]{hyperref}
\usepackage{amsmath}
\usepackage{amsfonts,amssymb,mathrsfs}
\usepackage{booktabs}
\usepackage{subfigure}

\usepackage{comment}

\usepackage{makecell}
\usepackage{multirow}
\usepackage{amsthm}
\usepackage{verbatim}
\usepackage{amsfonts,  amssymb,  mathrsfs}
\usepackage[subfigure]{tocloft}
\usepackage{subfigure}
\usepackage{xcolor}

\usepackage[ruled,linesnumbered]{algorithm2e}

\usepackage{ulem}

\ifCLASSINFOpdf
\else
\fi
\begin{document}
\bstctlcite{IEEEtran:BSTcontrol}
%
\title{End-to-end Conditional Diffusion for Realistic and Controllable Visual Traffic Scenario Generation}

%
%
%

\author{Jingzheng~Li,
        Yufei~Ge,
        Zhijun~Chen,
        Qianren~Mao, 
        Zizhe~Wang,
        Binhang~Qi,
        Bing~Li,
        Keyu~Chen,
        Baochang~Zhang,~\IEEEmembership{Member,~IEEE,}
        Xianglong~Liu${^*}$,~\IEEEmembership{Senior Member,~IEEE,}
        and Philip S. Yu,~\IEEEmembership{Life Fellow,~IEEE}
\thanks{Jingzheng Li and Qianren Mao are with Zhongguancun Laboratory, Beijing 100094, China. (email:maxlijingzheng@163.com)}
\thanks{Yufei Ge is with Tianjin University, Tianjin 300073, China.}
\thanks{Zhijun Chen, Zizhe Wang, Binhang Qi and Baochang Zhang are with Beihang University, Beijing 100091, China.}
\thanks{Bing Li is with Nanyang Technological University, 639789, Singapore.}
\thanks{Keyu Chen is with Tsinghua University, Beijing 100094, China.}
\thanks{Xianglong Liu is with the State Key Laboratory of Software Development Environment, Beihang University, Beijing 100191, China, and also with Zhongguancun Laboratory, Beijing 100094, China.}
\thanks{Philip S. Yu is with the University of Illinois Chicago, Chicago 60607, USA.}
\thanks{${^*}$ Corresponding author: Xianglong Liu (e-mail:xlliu@buaa.edu.cn).}
\thanks{Manuscript received MM, YY; revised MM, YY.}}

%
%

\markboth{Journal of \LaTeX\ Class Files,~Vol.~14, No.~8, August~2015}%
{Shell \MakeLowercase{\textit{et al.}}: Bare Demo of IEEEtran.cls for IEEE Journals}
%


\maketitle

\begin{abstract}
Generating closed-loop traffic scenarios that are both realistic and controllable is crucial for evaluating autonomous driving systems, especially under rare safety-critical interactions.
Existing learning-based methods often struggle to balance controllability and realism, offering either limited fine-grained control over traffic behavior or controllable scenarios at the expense of behavioral plausibility.
This paper presents E2E-CDiff, an end-to-end conditional diffusion framework for controllable and realistic scenario generation.
Conditioned on front-view visual observations, E2E-CDiff jointly denoises future motion states and executable low-level controls for route-interacting background vehicles. This unified state-action generation mitigates the planning-control mismatch in conventional two-stage trajectory-then-controller pipelines. Differentiable guidance further regulates speed, enforces drivable-area compliance, and supports collision-avoidance or collision-seeking behaviors, enabling both naturalistic and safety-critical scenario generation.
Experiments on Bench2Drive show that E2E-CDiff achieves a favorable controllability-realism trade-off compared with representative reinforcement- and imitation-learning baselines, while its collision-guided variant induces challenging interactions across multiple autonomous driving systems.
E2E-CDiff also performs competitively as a learning-based ego planner, demonstrating the generality of end-to-end state-action diffusion.

\end{abstract}

%
\IEEEpeerreviewmaketitle
\section{Introduction}
%
%
%
%
\IEEEPARstart{A}utonomous driving is progressing rapidly, yet rigorous safety evaluation remains difficult~\cite{feng2021intelligent,xu2022safebench}. 
Collecting diverse, safety‑critical real‑world data is costly, risky, and seldom covers long‑tail events~\cite{li2022coda}.
Closed‑loop simulation such as CARLA has therefore become central for assessing and training AD systems at scale.
Current simulation frameworks predominantly rely on hand-crafted scenarios, e.g., Bench2Drive~\cite{jiabench2drive}.
While such scenarios are useful for functional testing, they struggle to ensure realism and comprehensive coverage of real-world traffic interactions~\cite{Chen2025RIFT}.


Existing research has explored three main families of scenario generation methods~\cite{gao2025foundation}.
The first focuses on pixel-level scene synthesis using image- or video-based generative models~\cite{jiang2026dive}.
Although visually realistic, these scenarios lack physical interaction and do not readily support controllable closed-loop simulation, limiting their utility.
A second line of work leverages LLMs to produce scenario scripts such as OpenScenario or Scenic files~\cite{zhang2024chatscene,sheng2025talk2traffic}.
These methods excel in efficiency but typically require extensive prompt engineering and struggle to guarantee behavioral diversity or adherence to traffic rules.
The third comprises learning-based scenario generation methods that construct scenarios by adjusting the behaviors of traffic participants~\cite{hanselmann2022king,hao2023adversarial}.
Representative methods include adversarial reinforcement learning where surrounding vehicles are controlled as adversarial agents~\cite{chen2024frea}, and trajectory prediction based methods that manipulate agents' future motions~\cite{rempe2022generating,aiersilan2025generating}.
These methods have received considerable attention due to their ability to produce safety-critical, long-tail scenarios that are difficult to observe and collect in the real world.
However, existing learning-based scenario generation methods~\cite{sun2024drivescenegen} still suffer from limited controllability and insufficient realism.
These limitations highlight the need for a scenario generation method that can produce both controllable and realistic scenarios for reliable AD evaluation~\cite{Chen2025RIFT}.

To address these fundamental limitations, we propose E2E-CDiff, an End-to-End Conditional Diffusion Planner that simultaneously generates feasible trajectories and control signals through guided diffusion planning, enabling the construction of controllable and realistic scenarios.
E2E-CDiff introduces two key innovations: (1)\textbf{ end-to-end state-action control generation}, and (2) \textbf{adaptive conditional guidance}.
Specifically, traditional ADs, including existing diffusion-based approaches~\cite{liao2025diffusiondrive,xu2025diffscene}, generally adopt a two-stage architecture: they first generate high-level trajectory waypoints, then using low-level controllers (e.g., PID) to produce control signals. This decoupled design, however, creates a disconnect between trajectory planning and control execution, often leading to physically infeasible motions and suboptimal responsiveness in dynamic environments~\cite{wu2022trajectory}.
Thus, we design an end-to-end state-action generation mechanism within E2E-CDiff that jointly models future trajectories and executable controls in a unified diffusion process, reducing the planning-control mismatch and improving closed-loop execution.
On the other hand, we adopt an adaptive conditional guidance mechanism that steers the denoising phase of the diffusion process through gradients derived from differentiable objectives, including collision avoidance, speed limitation, and drivable-area adherence.
By parameterizing and weighting these objectives, the adaptive conditional guidance enables the flexible composition of control signals to satisfy diverse user-defined requirements, ranging from naturalistic, rule-compliant driving to safety-critical, adversarial interactions, thereby ensuring its \textbf{controllability}.

\begin{figure*}[t]
  \centering
\includegraphics[width=0.9\linewidth,scale=1]{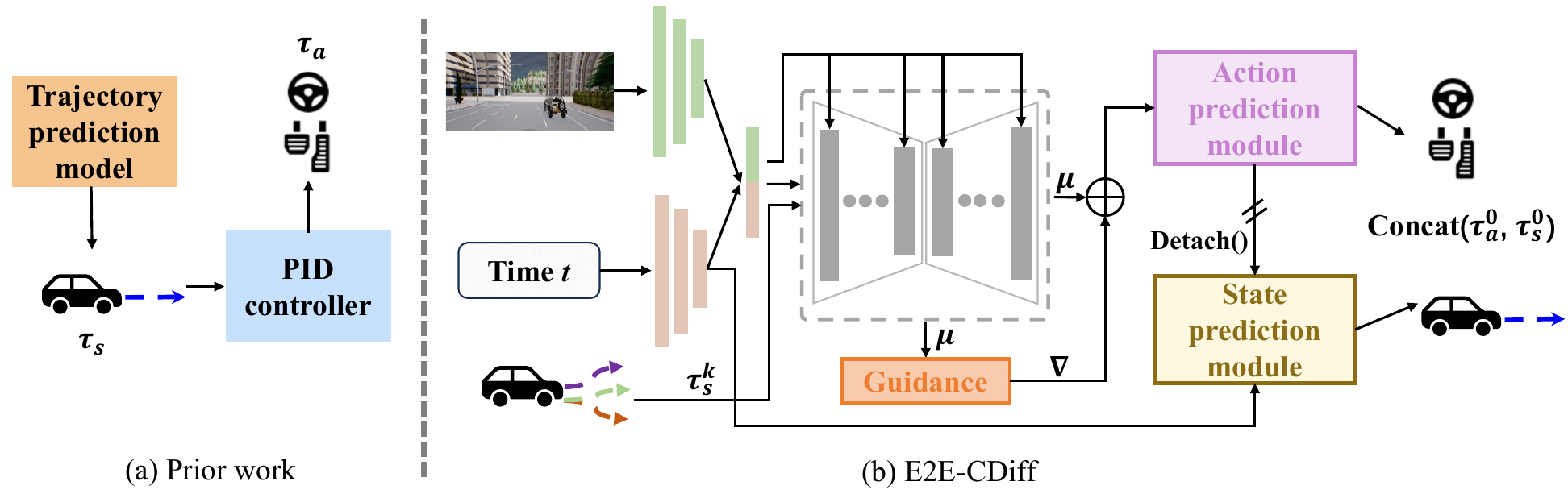}
\caption{Comparison between prior two-stage control and our end-to-end conditional diffusion planner. (a) Prior work predicts trajectories and then relies on a PID controller to obtain controls. (b) Our method denoises a state-action trajectory conditioned on front-view camera observations and diffusion step; during denoising, conditional guidance steers generation toward controllable yet naturalistic state-action behaviors. The model first predicts actions and then predicts states from detached action estimates before concatenating the two branches for joint rollout.}
  \label{framework1}
\end{figure*}

Our contributions can be summarized as follows:
\begin{itemize}

\item We design a learning-based scenario generation framework that enables realistic and controllable scenario construction.

\item We propose an end-to-end state-action diffusion planner that jointly generates future states and low-level controls to reduce the planning-control mismatch, together with an adaptive guidance mechanism for controllable scenario generation.

\item Experiments demonstrate that E2E-CDiff achieves a favorable trade-off between controllability and realism across different ego-vehicle planners, while its collision-guided variant generates challenging safety-critical scenarios for stress-testing AD systems.
\end{itemize}

\section{Related Work}
\subsection{Controllability and Realism in Traffic Simulation}

Effective evaluation of autonomous driving systems requires traffic simulators that provide both fine-grained controllability and high realism. Existing controllable simulation methods can be broadly grouped into three categories. The first category is conditional generative modeling, which learns a scene prior from data and steers generation at inference time through guidance signals, constraint functions, or language prompts, enabling control over scenario attributes such as speed profiles, reachability, and collision types. Representative examples include CTG~\cite{Zhong2023CTG}, SceneControl~\cite{Lu2024SceneControl}, and Scenario Diffusion~\cite{Pronovost2023ScenarioDiffusion}. The second category is retrieval-augmented generation, where realistic behavior fragments or exemplar scenarios are retrieved and then composed or edited to satisfy user intent while preserving motion fidelity, as in RealGen~\cite{Ding2024RealGen} and Chat2Scenario~\cite{zhao2024chat2scenario}. The third category is optimization-based scenario generation, e.g., CaDRE~\cite{Huang2024CaDRE}, AdaptiveStress~\cite{Koren2021AdaptiveStress} and DEEPScenario~\cite{Huang2022DEEPScenario}, which formulates scenario construction as a quality-diversity or black-box optimization problem under target control dimensions such as criticality, traffic density, or driving style.
Meanwhile, realism-oriented simulation frameworks, such as CARLA, Waymax~\cite{Gulino2024Waymax}, and nuPlan~\cite{Caesar2021nuPlan}, emphasize naturalistic traffic flow by leveraging reactive agents and real-world driving data. More recent generative simulators, including SceneGen~\cite{Tan2021SceneGen}, TrafficGen~\cite{feng2023trafficgen}, SceneDiffuser++~\cite{Tan2025SceneDiffuser}, and CCDiff~\cite{Lin2025CCDiff}, further improve interaction richness and long-horizon consistency, while hybrid training paradigms such as RIFT~\cite{Chen2025RIFT} and Ctrl-sim~\cite{Rowe2024CtRLSim} enhance behavioral adaptability in closed-loop settings. Nevertheless, existing methods still struggle to simultaneously achieve precise controllability, realistic interactive behavior, and robust closed-loop transfer, especially in safety-critical scenario generation. This gap motivates our end-to-end conditional diffusion framework, which aims to unify controllability and realism in scenario construction.
\subsection{Safety-critical Scenario Generation}
Unlike controllable and realistic traffic simulation~\cite{li2023scenarionet,wei2024editable}, safety-critical scenario generation focuses on constructing rare but hazardous interactions to efficiently evaluate the robustness of autonomous driving systems. Existing methods can be broadly categorized into three groups~\cite{ding2023survey}. 
Data-driven approaches~\cite{scanlon2021waymo} attempt to extract or synthesize risky scenarios from real-world driving datasets using generative models; however, their effectiveness is often limited by the severe imbalance between normal and safety-critical events. Adversarial-based methods~\cite{xu2025diffscene,hanselmann2022king}, which are currently the dominant paradigm, generate challenging scenarios by learning policies for surrounding agents or by directly perturbing environment states and transition dynamics to interfere with the ego vehicle’s decision-making process. These methods can be further divided into driving policy-based approaches~\cite{feng2021intelligent}, typically formulated as reinforcement learning problems~\cite{chen2021adversarial}, and transition function-based approaches~\cite{zhang2023cat}, which manipulate kinematic states or predicted trajectories to induce failures. Knowledge-based methods~\cite{cai2020summit} construct or validate safety-critical scenarios using domain knowledge such as traffic rules, accident statistics, and expert priors, offering stronger interpretability and better alignment with real-world safety concerns. In this work, we mainly focus on adversarial-based scenario generation due to its flexibility and effectiveness in producing challenging interactions across diverse traffic settings, while also considering knowledge-based safety-critical scenarios as complementary cases for functional evaluation.

\section{Methodology}
\label{sec:Methodology}
We first present the problem formulation and notation. We then introduce our end-to-end conditional diffusion framework for scenario generation. The framework consists of three components: a CBV identification module, which selects the background vehicle most likely to interact with the ego vehicle; an end-to-end conditional diffusion planner, which generates the CBV’s future states and actions; and a conditional guidance module, which directs the generation process toward user-specified objectives to ensure controllability while preserving realism.

\subsection{Problem Formulation}
\label{sec:problem_formulation}
We consider a simulated interactive traffic environment with $N$ agents, where one serves as the ego vehicle governed by the planner $\pi$, and the remaining $N-1$ are reactive agents represented by a background behavior model $\pi_{\mathrm{bg}}$. Our goal is to construct a closed-loop simulation in which reactive agents exhibit realistic and controllable driving behaviors. Specifically, among these $N-1$ agents, one or more are designated as critical background vehicles based on their potential interactions with the ego vehicle.


At any given timestep $t$, we denote the states of $N$ vehicles as $\mathbf{s}_t=[\mathbf{s}_t^1,\ldots,\mathbf{s}_t^N]$, where $\mathbf{s}_t^i=(x_t^i,y_t^i,\psi_t^i,v_t^i)$ denotes the 2D position, heading, and speed of vehicle $i$. The corresponding low-level control actions are $\mathbf{a}_t=[\mathbf{a}_t^1,\ldots,\mathbf{a}_t^N]$, where $\mathbf{a}_t^i=(\mathrm{throttle}_t^i,\mathrm{steer}_t^i,\mathrm{brake}_t^i)$.
Given the current state and action, the simulator advances the vehicle state as $\mathbf{s}_{t+1}^i=F_{\text{sim}}(\mathbf{s}_t^i,\mathbf{a}_t^i)$. In addition, we denote $\mathbf{c}_t^i$ as decision-relevant context for vehicle $i$, such as vehicle-centric map information or a front-view image, which helps continuously generate and update states and actions in closed-loop traffic simulation. For a selected CBV at the current decision cycle, we denote this conditioning context compactly by $\mathbf{c}$.


\subsection{Critical Background Vehicle Identification}
To address the rarity of safety-critical interactions, we intervene only on a small subset of background vehicles that are most likely to affect the ego vehicle, while keeping non-critical vehicles under rule-based control to maintain efficiency. 
We decouple interaction selection from behavior generation: for each surrounding vehicle, we estimate a route-level interaction score relative to the ego’s planned global route by evaluating potential future conflict points and their temporal consistency. 
Vehicles with the highest scores are selected as critical background vehicles (CBVs). 
For each selected CBV, we further synthesize a conflict-aware route anchored at the predicted conflict point and constrained by lane connectivity and drivability, providing an AV-centered initialization for subsequent closed-loop scenario generation.

\subsection{End-to-End State-Action Diffusion Planning}
As shown in Fig.~\ref{framework1}, prior two-stage methods first predict geometric trajectories and then convert them into control signals using an external PID controller. 
This decoupled pipeline can introduce planning-control mismatch in interactive closed-loop simulation. 
In contrast, our method unifies planning and control within a single  diffusion model that directly denoises a future state-action trajectory conditioned on context $\mathbf{c}$.
At each decision cycle, we start with Gaussian noise and iteratively denoise it into a clean rollout over a horizon of $T$ future steps.
\subsubsection{State-Action Trajectory Representation}
For a selected CBV, we omit the vehicle index and denote its clean state-action trajectory as $\boldsymbol{\tau}^{0}$, which is decomposed into a state sequence $\boldsymbol{\tau}_s$ and an action sequence $\boldsymbol{\tau}_a$:
\begin{equation}
\begin{aligned}
\boldsymbol{\tau}^{0}:=\begin{bmatrix}\boldsymbol{\tau}_s\\\boldsymbol{\tau}_a\end{bmatrix},
\boldsymbol{\tau}_a:=[\mathbf{a}_0, \mathbf{a}_1, \ldots, \mathbf{a}_{T-1}],\quad
\boldsymbol{\tau}_s:=[\mathbf{s}_1, \mathbf{s}_2, \ldots, \mathbf{s}_T].
\end{aligned}
\end{equation}
where $T$ is the prediction horizon, i.e., the number of future time steps in one rollout. Thus, $\boldsymbol{\tau}_a$ contains $T$ control actions and $\boldsymbol{\tau}_s$ contains $T$ future states. In addition, $\boldsymbol{\tau}_s \in \mathbb{R}^{T \times d_s}$ and $\boldsymbol{\tau}_a \in \mathbb{R}^{T \times d_a}$ follow the state/action definitions in Sec.~\ref{sec:problem_formulation}, with $d_s=4$ and $d_a=3$, respectively. Therefore, the total trajectory dimension is $d=d_s+d_a=7$. In this paper, ``trajectory'' refers to this state-action trajectory.

\subsubsection{End-to-End Architecture Design}
At diffusion step $k\in\{1,\ldots,K\}$ (with $K$ the total number of denoising steps), let $\boldsymbol{\tau}^{k}$ denote the noisy state-action trajectory. We adopt a clean-trajectory ($\boldsymbol{\tau}^{0}$) prediction parameterization with a decoupled action-state head:
\begin{equation}
\hat{\boldsymbol{\tau}}_{a,\theta}^{0}=f_{\text{act}}(\boldsymbol{\tau}^{k},k,\mathbf{c}),\quad
\hat{\boldsymbol{\tau}}_{s,\theta}^{0}=f_{\text{state}}(\operatorname{detach}(\hat{\boldsymbol{\tau}}_{a,\theta}^{0}),k,\mathbf{c}).
\end{equation}

The action head predicts clean low-level controls, while the state head predicts the corresponding clean state rollout using actions detached from the computational graph, thereby avoiding unstable gradient coupling in standard denoising. During guided control optimization, we recompute the states from the guided actions without detaching them, to preserve gradients to the control sequence. The branch outputs are stacked as $\hat{\boldsymbol{\tau}}_{\theta}^{0}(\boldsymbol{\tau}^{k},k,\mathbf{c})=[\hat{\boldsymbol{\tau}}_{s,\theta}^{0};\hat{\boldsymbol{\tau}}_{a,\theta}^{0}]$, following the same state-action order as $\boldsymbol{\tau}^{0}=\begin{bmatrix}\boldsymbol{\tau}_s\\\boldsymbol{\tau}_a\end{bmatrix}$, and are used to parameterize the reverse transition in Eq.~\eqref{eq4}. This design jointly models motion evolution and control generation within a single network and directly provides executable control signals $(\mathrm{throttle},\mathrm{steering},\mathrm{braking})$ in closed-loop simulation.

\subsubsection{Forward Diffusion Process}
Let $\boldsymbol{\tau}^{k}$ be the state-action trajectory at the $k$-th diffusion step and $\boldsymbol{\tau}^{0}$ be the clean trajectory. The forward diffusion process gradually corrupts the clean trajectory $\boldsymbol{\tau}^{0}$ over $K$ steps by adding Gaussian noise:
\begin{equation}
\begin{aligned}
&q(\boldsymbol{\tau}^{1:K}|\boldsymbol{\tau}^{0}):=\prod_{k=1}^{K}q(\boldsymbol{\tau}^{k}|\boldsymbol{\tau}^{k-1})\\
&q(\boldsymbol{\tau}^{k}|\boldsymbol{\tau}^{k-1}):=\mathcal{N}(\boldsymbol{\tau}^{k};\sqrt{1-\beta_{k}}\boldsymbol{\tau}^{k-1},\beta_{k}\mathbf{I}),
\end{aligned}
\end{equation}
where $\{\beta_k\}_{k=1}^K$ is a fixed variance schedule. The closed-form expression after $k$ steps is:
\begin{equation}
\boldsymbol{\tau}^{k} = \sqrt{\bar\alpha_k}\,\boldsymbol{\tau}^{0} + \sqrt{1-\bar\alpha_k}\,\boldsymbol{\varepsilon}, \quad
\boldsymbol{\varepsilon} \sim \mathcal{N}(\mathbf{0}, \mathbf{I}),
\end{equation}
with $\bar\alpha_k = \prod_{i=1}^{k}(1 - \beta_i)$.

\subsubsection{Reverse Diffusion Process}
Given a noisy trajectory $\boldsymbol{\tau}^{k}$ and context information $\mathbf{c}$, the reverse process is:
\begin{equation}
\begin{aligned}
p_{\theta}(\boldsymbol{\tau}^{0:K}|\mathbf{c})&:=p(\boldsymbol{\tau}^{K})\prod_{k=1}^{K}p_{\theta}(\boldsymbol{\tau}^{k-1}|\boldsymbol{\tau}^{k},\mathbf{c})\\p_{\theta}(\boldsymbol{\tau}^{k-1}|\boldsymbol{\tau}^{k},\mathbf{c})&:=\mathcal{N}(\boldsymbol{\tau}^{k-1};\boldsymbol{\mu}_{\theta}(\boldsymbol{\tau}^{k},k,\mathbf{c}),\boldsymbol{\Sigma}_{\theta}(\boldsymbol{\tau}^{k},k,\mathbf{c})),
\label{eq4}
\end{aligned}
\end{equation}
The denoiser predicts the clean trajectory $\hat{\boldsymbol{\tau}}_{\theta}^{0}$, from which the reverse mean $\boldsymbol{\mu}_{\theta}$ is computed using the standard DDPM posterior parameterization~\cite{ho2020denoising}. Here, $\theta$ denotes the parameters of the diffusion model and the variance term of the Gaussian
transition is fixed as $\boldsymbol{\Sigma}^{k}=\boldsymbol{\Sigma}_{\theta}(\boldsymbol{\tau}^{k},k,\mathbf{c})=\sigma_{k}^{2}\mathbf{I}=\beta_{k}\mathbf{I}$. 
Trajectory samples are generated by iteratively applying:
\begin{equation}
\boldsymbol{\tau}^{k-1} = \boldsymbol{\mu}_{\theta}(\boldsymbol{\tau}^{k}, k,\mathbf{c}) + \sigma_k \boldsymbol{z}, \quad \boldsymbol{z} \sim \mathcal{N}(\mathbf{0}, \mathbf{I}),
\end{equation}
starting from $\boldsymbol{\tau}^K \sim \mathcal{N}(\mathbf{0}, \mathbf{I})$.

\subsubsection{Training Objective}
The model parameters $\theta$ are optimized to reverse the noising process. A simplified loss function minimizes the discrepancy between the clean trajectory and the model’s prediction:
\begin{equation}
\mathcal{L}(\theta) = 
\mathbb{E}_{\boldsymbol{\tau}^{0}, \boldsymbol{\varepsilon}, k, \mathbf{c}}\big[
\lVert \boldsymbol{\tau}^{0} - \hat{\boldsymbol{\tau}}_{\theta}^{0}(\boldsymbol{\tau}^{k}, k,\mathbf{c}) \rVert_2^2
\big],
\end{equation}
where $\boldsymbol{\tau}^{k}$ is obtained via the forward process at diffusion step $k$.    

\subsection{Conditional Guidance for Scenario Generation}
\begin{figure}[t]
  \centering
\includegraphics[width=0.99\linewidth,scale=1]{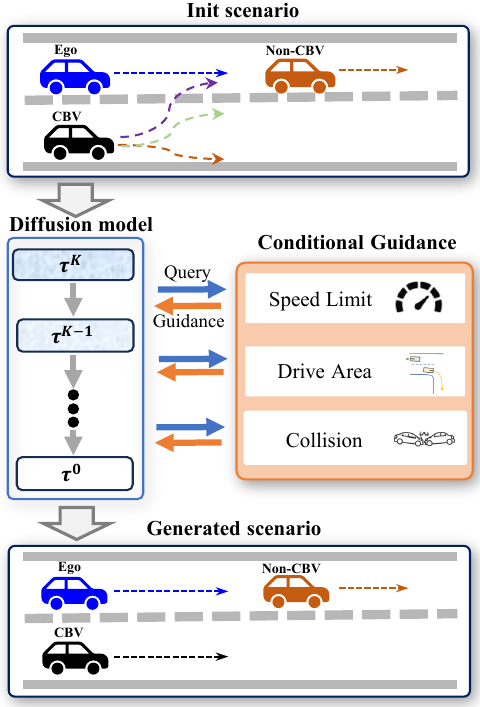}
  \caption{Conditional guidance in reverse diffusion. At each denoising step, objective gradients modify the mean update and steer sampled trajectories toward user-specified constraints.}
  \label{framework2}
\end{figure}

Our diffusion model learns realistic state-action trajectories for traffic scenario generation.
To make the generated scenarios controllable, we incorporate conditional guidance into reverse diffusion, as illustrated in Fig.~\ref{framework2}, so that sampled trajectories satisfy predefined objectives (e.g., rule compliance or risk-oriented behaviors).
Let $\mathcal{J}$ denote the objective encoding desired scenario constraints.  
At each reverse diffusion step, we guide sampling by shifting the Gaussian mean along the objective gradient:
\begin{equation}
p_{\theta}(\boldsymbol{\tau}^{k-1}\mid\boldsymbol{\tau}^{k},\mathbf{c})
\approx
\mathcal{N}\!\left(
\boldsymbol{\tau}^{k-1};
\boldsymbol{\mu}_{\theta}(\boldsymbol{\tau}^{k},k,\mathbf{c})-\boldsymbol{\Sigma}^{k}\mathbf{g}^{k},
\boldsymbol{\Sigma}^{k}
\right),
\end{equation}
where $\mathbf{g}^{k}=\nabla_{\boldsymbol{\tau}^{k}}\mathcal{J}\!\left(\hat{\boldsymbol{\tau}}_{\theta}^{0}(\boldsymbol{\tau}^{k},k,\mathbf{c})\right)$ denotes the gradient of the objective evaluated on the predicted clean trajectory at diffusion step $k$.  
This guided reverse process iteratively improves controllability while preserving trajectory realism.
Below we instantiate several guidance objectives for controllable scenario generation.

\subsubsection{Controlled Guidance Generation under Basic Driving Conditions} 
\paragraph{Speed Guidance} Here, we consider the speed-related constraint focusing on controlling
the speed of the CBV.
We penalize deviations from a desired speed profile. Let \(v_t^{\mathrm{cbv}}\) be the CBV speed at time step \(t\) and \(v^*\) the target speed. We formulate the objective as
\begin{equation}
\mathcal{J}_{\mathrm{speed}}(\boldsymbol{\tau}^{0})=
\sum_{t=1}^{T} \|v_t^{\mathrm{cbv}} - v^*\|_2^2.
\label{eq:speed_loss}
\end{equation}
By minimizing this objective, the CBV speed is encouraged to approach the target speed \(v^*\).
\paragraph{Drivable Area Guidance} This constraint encourages the generated trajectory to remain within the drivable area. Let $\mathcal{A}$ represent the drivable area, and let $d_{\mathcal{A}}(\mathbf{p})$ denote the signed distance from position $\mathbf{p}$ to its boundary, defined as positive inside $\mathcal{A}$, zero on the boundary, and negative outside $\mathcal{A}$. Here, $\mathbf{p}_t^{\mathrm{cbv}}$ is the CBV position at time step $t$.
The drivable area constraint is formulated as:
\begin{equation}
\mathcal{J}_{\mathrm{area}}(\boldsymbol{\tau}^{0}) = \sum_{t=1}^{T} \max\!\left(0, \delta_{\mathrm{margin}} - d_{\mathcal{A}}(\mathbf{p}_t^{\mathrm{cbv}})\right)^2,
\label{eq:area_loss}
\end{equation}
where $\delta_{\mathrm{margin}}$ is a safety margin. The penalty activates near the boundary and increases when the CBV moves outside the drivable area because the signed distance becomes negative.
\paragraph{Collision Guidance}
To generate naturalistic scenarios, we first constrain the CBV to avoid collisions with the ego vehicle.
Let $\mathbf{p}_t^{\mathrm{ego}}$ and $\mathbf{p}_t^{\mathrm{cbv}}$ denote the ego and CBV positions at time step $t$, and define
$d_t=\|\mathbf{p}_t^{\mathrm{ego}}-\mathbf{p}_t^{\mathrm{cbv}}\|_2$.
The collision-avoidance loss is:
\begin{equation}
\mathcal{J}_{\mathrm{coll}}^{\mathrm{safe}}(\boldsymbol{\tau}^{0})
=
\sum_{t=1}^{T}
\left[
\operatorname{softplus}\!\left(\frac{d_t^{\mathrm{safe}}-d_t}{\kappa}\right)
+\lambda\exp\left(-\frac{d_t^2}{2\sigma^2}\right)
\right],
\end{equation}
where $d_t^{\mathrm{safe}}$ is the safety distance determined by vehicle size and a buffer margin, and $\kappa,\lambda,\sigma$ control smoothness and distance shaping. Minimizing $\mathcal{J}_{\mathrm{coll}}^{\mathrm{safe}}$ penalizes close interactions and prevents collisions.

To construct safety-critical scenarios, we enable collision-seeking guidance only when the ego--CBV distance falls within a trigger range. Let $d_{\mathrm{ref}}$ be the current ego--CBV distance and define a trigger gate $\eta(d_{\mathrm{ref}})\in[0,1]$ (hard or smooth), where $\eta\!\approx\!1$ for $d_{\mathrm{ref}}\le d_{\mathrm{th}}$ and $\eta\!\approx\!0$ otherwise. The risk-seeking loss is:
\begin{equation}
\mathcal{J}_{\mathrm{coll}}^{\mathrm{risk}}(\boldsymbol{\tau}^{0})
=
-\eta(d_{\mathrm{ref}})\,\mathcal{J}_{\mathrm{coll}}^{\mathrm{safe}}(\boldsymbol{\tau}^{0}).
\end{equation}
Minimizing $\mathcal{J}_{\mathrm{coll}}^{\mathrm{risk}}$ encourages closer interaction only in the triggered distance range, which helps generate safety-critical scenarios without globally forcing aggressive behavior.


\subsubsection{Multi-Objective Guidance Framework}
\label{sec3.4.2}
In practice, multiple guidance objectives need to be combined to generate scenarios that are both realistic and controllable and that meet specific testing requirements. 
We propose a weighted combination framework that allows flexible control over different guidance mechanisms.
At each reverse step, we combine multiple guidance objectives through a weighted sum. We instantiate the total guidance objective according to scenario type:

\begin{itemize}
\item \textbf{Naturalistic Driving Scenarios}: use collision-avoidance guidance
\begin{equation}
\mathcal{J}_{\mathrm{total}}^{\mathrm{nat}}=
w_{\mathrm{speed}}\,\mathcal{J}_{\mathrm{speed}}
+w_{\mathrm{area}}\,\mathcal{J}_{\mathrm{area}}
+w_{\mathrm{coll}}\,\mathcal{J}_{\mathrm{coll}}^{\mathrm{safe}},
\end{equation}
where $w_{\mathrm{speed}}$, $w_{\mathrm{area}}$, and $w_{\mathrm{coll}}$ are the weights for speed, drivable-area, and collision guidance, respectively.
\item \textbf{Safety-Critical Scenarios}: switch to collision-seeking guidance
\begin{equation}
\mathcal{J}_{\mathrm{total}}^{\mathrm{sc}}=
w_{\mathrm{speed}}\,\mathcal{J}_{\mathrm{speed}}
+w_{\mathrm{area}}\,\mathcal{J}_{\mathrm{area}}
+w_{\mathrm{coll}}\,\mathcal{J}_{\mathrm{coll}}^{\mathrm{risk}},
\label{eq14}
\end{equation}
with $\mathcal{J}_{\mathrm{coll}}^{\mathrm{risk}}(\boldsymbol{\tau}^{0})=-\eta(d_{\mathrm{ref}})\mathcal{J}_{\mathrm{coll}}^{\mathrm{safe}}(\boldsymbol{\tau}^{0})$, to generate safety-critical interactions in the triggered distance range.
\end{itemize}
\subsection{Implementation details}
The context $\mathbf{c}$ is represented as a vehicle-centric front-view camera image and encoded by a ResNet-34 backbone.  
Similar to Diffuser~\cite{janner2022diffuser}, we use a diffusion model
architecture like U-Net containing several blocks of temporal 1D convolutional blocks and skip connections over the input trajectory.
The timestep embedding is produced by a sinusoidal positional encoder followed by an MLP, then concatenated with the image features and injected into every residual block.
The denoiser factorizes its predictions into an action branch and a dedicated state predictor: an action head predicts 3-D controls $(\mathrm{throttle},\mathrm{steering}, and \mathrm{braking})$, and a state head predicts the 4-D state rollout from the action sequence and time embedding. The two branches are concatenated to form a 7-D clean state-action trajectory prediction. We use a squared-cosine variance schedule with $K=100$ diffusion steps, and keep $K=100$ denoising steps in inference.

\section{Experiments}
\begin{figure*}[t]
  \centering
\includegraphics[width=0.99\linewidth,scale=1]{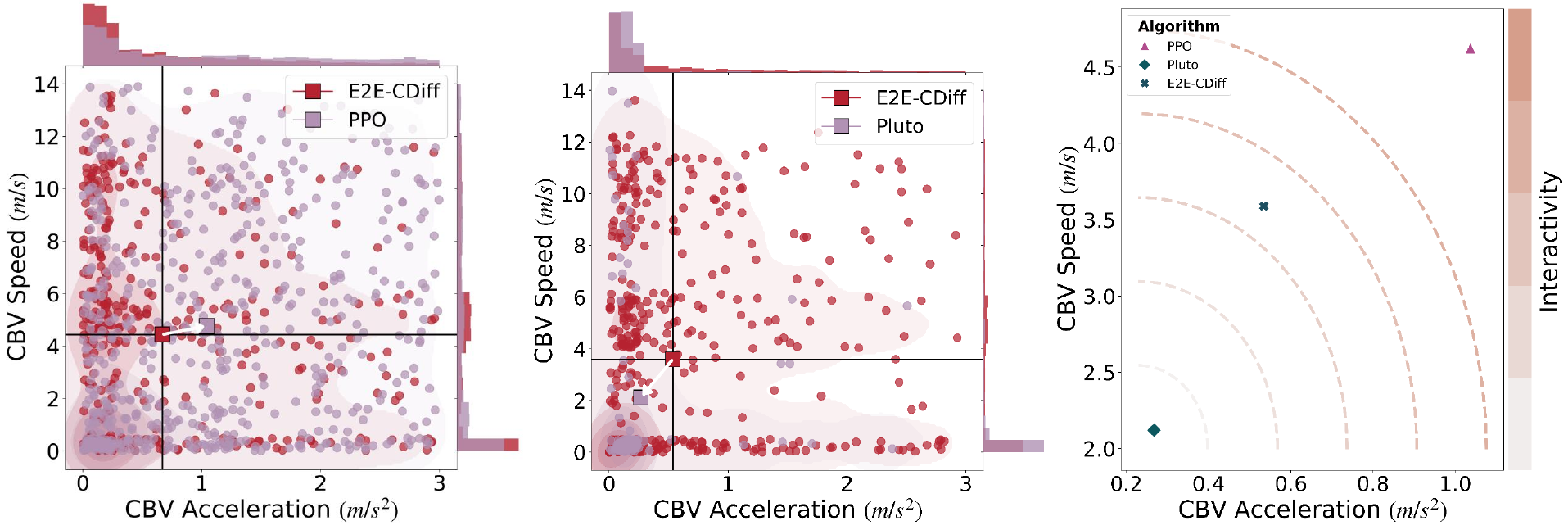}
  \caption{Speed and acceleration distribution of CBVs.}
  \label{distribution}
\end{figure*}
\begin{figure}[t]
  \centering
\includegraphics[width=0.9\linewidth,scale=1]{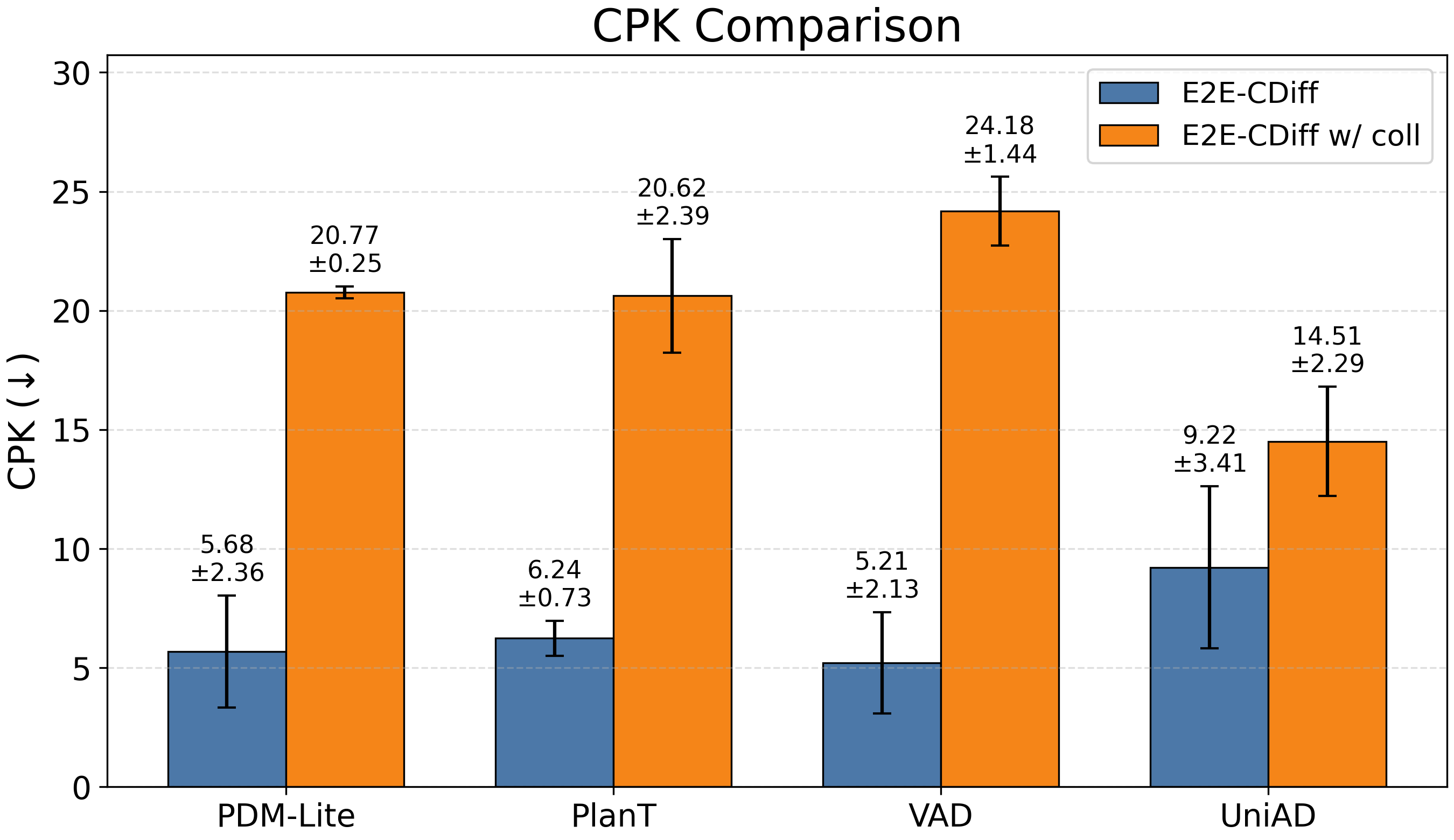}
  \caption{CPK comparison between E2E-CDiff and E2E-CDiff w/ coll.}
  \label{cpk}
\end{figure}
\begin{table}[t]
\Large
\caption{Controllability and realism under PDM-Lite ego planner.}
\centering
{\resizebox{\linewidth}{!}{
\begin{tabular}{cccccc}
\toprule
\multirow{2}{*}{Methods}&\multicolumn{2}{c}{Controllability}&\multicolumn{3}{c}{Realism}\\
\cmidrule{2-3} \cmidrule{4-6}
&2D-TTC$\uparrow$ &ORR$\downarrow$&S-SW$\uparrow$&A-SW$\uparrow$&S-WD$\downarrow$\\
\midrule
PPO&2.41$\pm$1.52&8.82$\pm$8.83&0.80$\pm$0.01&0.65$\pm$0.02&6.48$\pm$0.10\\
Pluto&2.56$\pm$1.55&3.02$\pm$3.09 &0.84$\pm$0.08 &0.83$\pm$0.12&6.13$\pm$0.59\\
RIFT&2.70$\pm$1.34&\textbf{0.83$\pm$0.33}&0.97$\pm$0.00&\textbf{0.94$\pm$0.03}&\textbf{4.26$\pm$0.12}\\
E2E-CDiff&\textbf{2.79$\pm$1.33}&1.26$\pm$1.47&\textbf{0.97$\pm$0.01}&0.88$\pm$0.09&4.80$\pm$1.38\\
\bottomrule
\end{tabular}
}}
\label{tab1}
\end{table}
\begin{table}[t]
\Large
\caption{Controllability and realism under  PlanT ego planner.}
\centering
{\resizebox{\linewidth}{!}{
\begin{tabular}{cccccc}
\toprule
\multirow{2}{*}{Methods}&\multicolumn{2}{c}{Controllability}&\multicolumn{3}{c}{Realism}\\
\cmidrule{2-3} \cmidrule{4-6}
&2D-TTC$\uparrow$ &ORR$\downarrow$&S-SW$\uparrow$&A-SW$\uparrow$&S-WD$\downarrow$\\
\midrule
PPO&2.04$\pm$2.08&4.50$\pm$1.15&0.93$\pm$0.01&0.94$\pm$0.02&5.02$\pm$0.19\\
Pluto&2.59$\pm$1.88&6.44$\pm$1.57&0.84$\pm$0.02&0.78$\pm$0.03&6.20$\pm$0.23\\
RIFT&1.90$\pm$0.36&\textbf{0.38$\pm$0.10}&0.63$\pm$0.03&0.49$\pm$0.02&7.23$\pm$0.08\\
E2E-CDiff&\textbf{2.92$\pm$1.53}&5.96$\pm$2.47&\textbf{0.97$\pm$0.03}&\textbf{0.94$\pm$0.02}&\textbf{3.83$\pm$0.73}\\
\bottomrule
\end{tabular}
}}
\label{tab2}
\end{table}

We conducted experiments to validate the following research questions.
\textbf{RQ1}: Does end-to-end conditional diffusion model generate scenarios with better realism and controllability compared to other driving planners?
\textbf{RQ2}: Can collision-oriented guidance generate challenging safety-critical scenarios across different autonomous driving systems?
\textbf{RQ3}: What is the impact of different components of E2E-CDiff on the overall results?
\textbf{RQ4}: Beyond controlling CBVs to construct traffic flow, how effective is E2E-CDiff when used as the ego-vehicle planner?


\subsection{Experiment Setups}

\subsubsection{Baselines and Dataset}
\paragraph{Baselines}
To evaluate the effectiveness of scenario generated via diffusion models, we employ various autonomous driving algorithms as the planners for the ego vehicle (ego planners), including rule-based, imitation learning-based, and end-to-end algorithms. 
Meanwhile, different autonomous driving algorithms are utilized to control the CBV  (CBV planner) to generate scenario as baselines.
Except for the proposed diffusion model, we also compare reinforcement learning-based PPO, imitation learning-based PlanT.

\textbf{Ego Planner:}
(1) \textbf{PDM-Lite}~\cite{Beißwenger2024PdmLite}: A rule-based privileged expert method that achieves state-of-the-art performance on the CARLA Leaderboard 2.0 by leveraging components such as the Intelligent Driver Model and
the kinematic bicycle model.
(2) \textbf{PlanT}~\cite{renz2022plant}: An explainable, learning-based planning method that operates on an object-level input representation and is trained through imitation learning.
(3) \textbf{UniAD}~\cite{hu2023planning}: A planning-oriented unified framework integrating perception, prediction, mapping, and planning into one end-to-end model using query-based interfaces.
(4) \textbf{VAD}~\cite{jiang2023vad}: A fast, end-to-end vectorized driving paradigm representing scenes with vectorized motion and map elements for efficient, safe planning.

\textbf{CBV Planner:}
(1) \textbf{Pluto}~\cite{cheng2024pluto} is an open-source IL-based planning framework for autonomous driving. It processes
vectorized scene representations as input and outputs multimodal trajectories for downstream planning. In the AV-centric closed-loop simulation, the method directly uses a pre-trained checkpoint without additional fine-tuning.
(2) \textbf{PPO}~\cite{schulman2017proximal} is a widely used reinforcement learning algorithm known for its balance between simplicity and performance.
(3) \textbf{RIFT}~\cite{Chen2025RIFT} is a dual-stage traffic simulation framework that combines imitation learning pre-training with group-relative reinforcement learning fine-tuning to improve realism and controllability in closed-loop simulation.
\paragraph{Dataset}
In experiments, we use the Bench2Drive Benchmark based on CARLA version 0.9.15. We train the Ego planner and CBV planner using training data comprising 220 routes. During the testing phase, we employ 10 routes to evaluate scenario generation ability and driving performance.
\subsubsection{Metrics}
We evaluate our method from two complementary aspects: \textit{controllability} and \textit{realism}, aiming to verify that the proposed E2E-CDiff can generate naturalistic driving scenarios that are both realistic and controllable.

\paragraph{Controllability Evaluation}  
To assess controllability, we focus on parameters directly affected by our guidance objectives, including relative speed and time-to-collision (TTC) between the ego and the CBV.
Following the metric design principles of the WOSAC challenge and prior studies, we adopt two standard metrics:

\textbf{2D Time-to-Collision (2D-TTC):} the minimum of longitudinal and lateral TTC from the ego vehicle’s perspective, capturing the interaction risk and proximity between the ego and CBV.

\textbf{Off-Road Rate (ORR)}: the percentage of time CBVs spend off-road on average.
\paragraph{Realism Evaluation}  
Unlike TrajData~\cite{ivanovic2023trajdata}, which contains real trajectories for statistical distribution comparisons, CARLA lacks expert demonstrations.
Thus, we adopt distribution-level measures to evaluate whether the generated CBV motions align with realistic speed and acceleration patterns:

\textbf{Speed Shapiro–Wilk Test (S-SW)} and \textbf{Acceleration Shapiro-Wilk Test (A-SW)}~\cite{shaphiro1965analysis}: evaluate the normality of speed and acceleration distributions, an assumption supported by empirical traffic studies, to capture statistical naturalness.

\textbf{Wasserstein Distance on Speed (S-WD)}~\cite{vaserstein1969markov}: measures the distributional distance between simulated and target speed distributions, quantifying the realism of motion behavior.
\paragraph{Safety-Critical Evaluation}
To assess the effectiveness of our collision-guided generation in producing challenging safety-critical scenarios, we compute:

\textbf{Driving Score (DS):} a composite performance score reflecting rule compliance, smoothness, and safety under adversarial conditions.

\textbf{Route Completion (RC):} the percentage of the route completed by an agent before it gets blocked or deviates from the route.

\textbf{Collision Per Kilometer (CPK):} the number of CBV collision events normalized by traveled distance (reported as collisions per kilometer), where lower values indicate safer interaction behavior.

\begin{figure*}[t]
  \centering
\includegraphics[width=0.99\linewidth,scale=1]{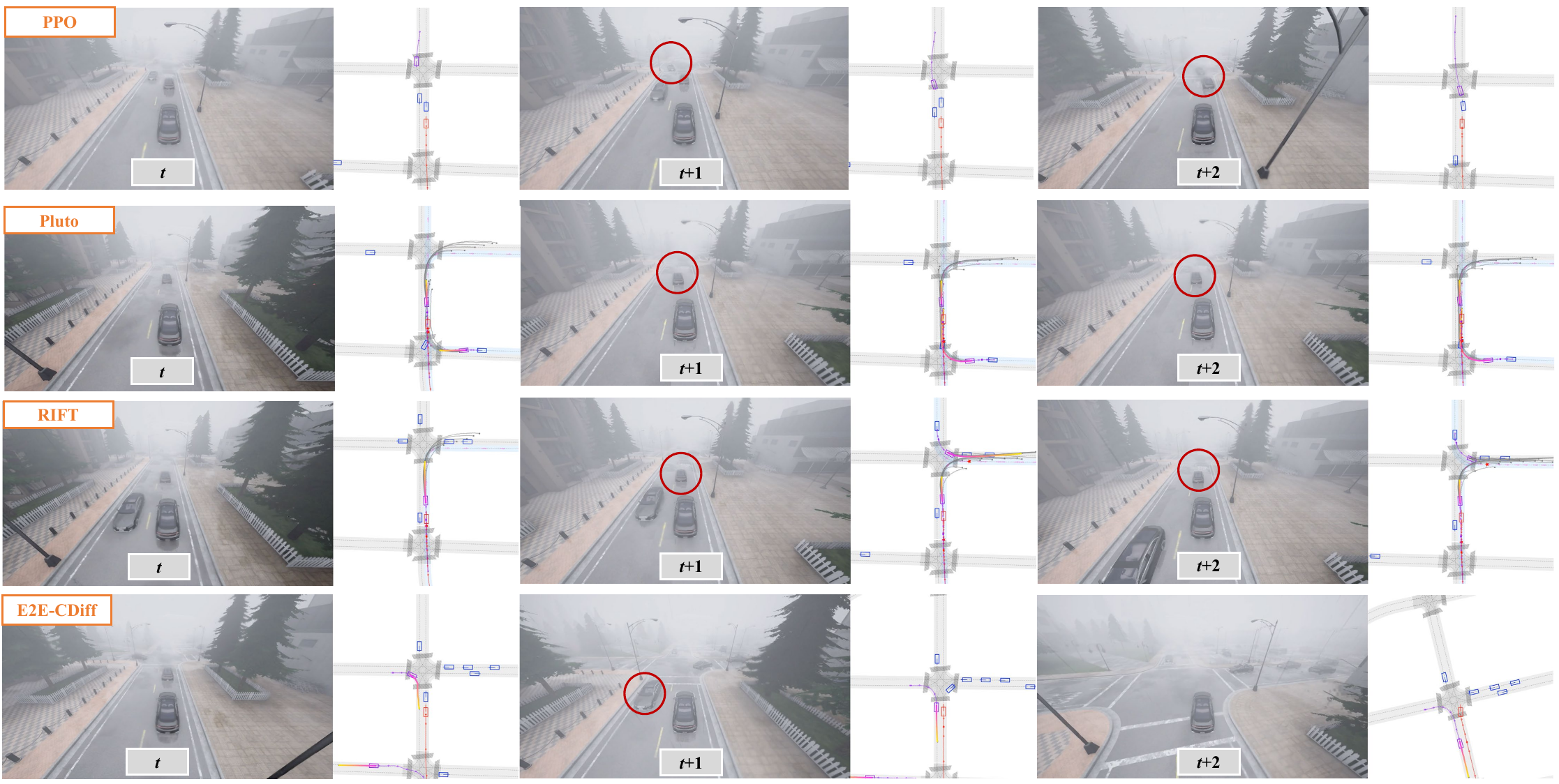}
  \caption{Qualitative comparison across CBV planners. For each planner, three temporal frames are shown. In each frame, the CBV is marked in purple, the ego vehicle (PDM-Lite) in red, and background vehicles in blue. Key interaction regions are highlighted with red circles.}
  \label{video}
\end{figure*}
\begin{figure*}[t]
  \centering
\includegraphics[width=0.99\linewidth,scale=1]{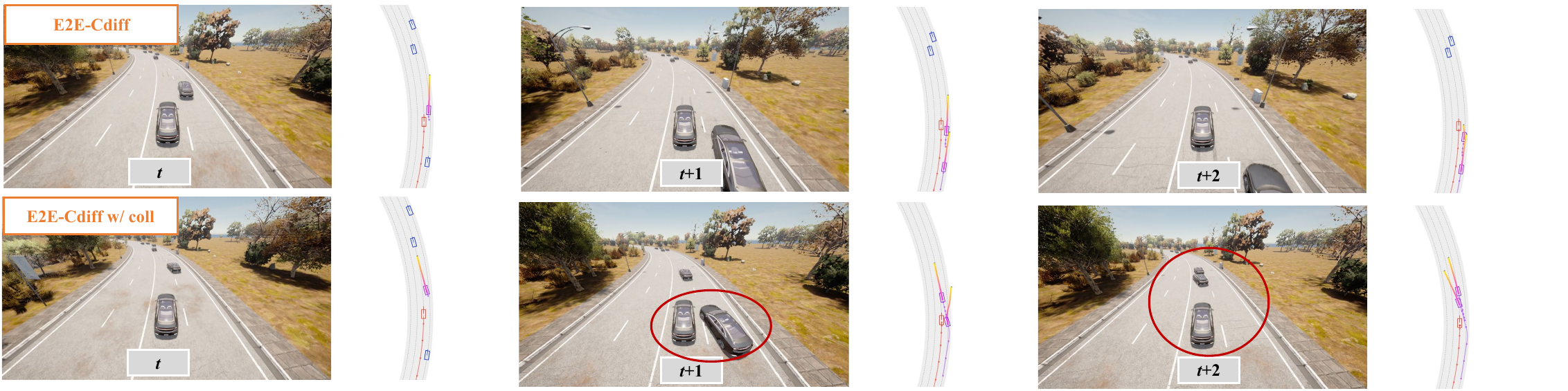}
  \caption{Qualitative results of safety-critical scenario generation between E2E-CDiff and E2E-CDiff w/ coll.}
  \label{video2}
\end{figure*}

\subsection{Realistic and Controllable Traffic Scenario Generation}

To address \textbf{RQ 1}, we evaluate the controllability and naturalness of scenario generated by different CBV controllers.
Tables \ref{tab1} and \ref{tab2} report the performance of generated scenario for different baselines when the AD system of ego vehicle is set to PDM-Lite and PlanT, respectively.
Overall, the scenarios generated by E2E-CDiff demonstrate consistently better controllability and realism than those produced by PPO and Pluto across different ego planners.

Furthermore, Fig.~\ref{distribution} visualizes the distribution of speed and acceleration for CBV controlled by different planners.
The PPO achieves higher speeds and accelerations, resulting in more interactions, while the Pluto maintains relatively lower speeds and accelerations, ensuring safer operation. Our E2E-CDiff strikes a balance between interaction and safety.                            

\subsection{Safety-Critical Scenario Generation}
To address \textbf{RQ2}, we conduct a safety-critical evaluation of various ego AD systems using scenarios generated by different CBV planners.
%
Specifically, we consider four ego AD systems, namely PDM-Lite, PlanT, UniAD, and VAD. The CBV planners evaluated in Table~\ref{tablescenario1} include three baselines, PPO, Pluto, and RIFT, and our proposed E2E-CDiff.
E2E-CDiff w/ coll denotes the collision-guided variant, which encourages collision-prone interactions between the CBV and the ego vehicle for safety-critical scenario generation and is evaluated separately in Fig.~\ref{cpk} and Fig.~\ref{video2}.
For details, please refer to Sec.~\ref{sec3.4.2}.

Table~\ref{tablescenario1} reports the driving score and route completion of each ego vehicle under scenarios generated by different CBV planners.
Experimental results demonstrate that for different CBV controllers, scenarios generated by E2E-CDiff achieve higher driving scores across different AD system of ego vehicle, indicating that E2E-CDiff can produce more realistic and controllable driving scenarios. 
PDM-Lite achieves the best driving score, because it is rule-based privileged expert method.
In addition, it can be seen that end-to-end approaches hold considerable potential as AD systems.


Furthermore, to evaluate collision-prone safety-critical scenario generation, we further compare the CPK of E2E-CDiff and E2E-CDiff w/ coll.
As shown in Fig.~\ref{cpk}, adding collision-oriented guidance consistently increases the CPK across different ego AD systems.
For example, the CPK increases from $5.68$ to $20.77$ for PDM-Lite.
Since CPK measures the number of collision events per kilometer, these results show that collision-oriented guidance produces more collision-prone interactions across different ego AD systems.

Fig.~\ref{video} provides a qualitative comparison of scenarios generated by different CBV planners, where the ego vehicle is controlled by PDM-Lite and the CBV is controlled by PPO, Pluto, RIFT, or E2E-CDiff.
For each planner, three temporal frames are presented to show the evolution of the interaction.
As highlighted in the figure, PPO drives the CBV into the oncoming lane, while Pluto and RIFT produce lane-boundary violations during right turns.
In contrast, the CBV controlled by E2E-CDiff follows the drivable lane more smoothly and exhibits more natural behavior without obvious traffic-rule violations.
Fig.~\ref{video2} further compares E2E-CDiff with its collision-guided variant, E2E-CDiff w/ coll, to illustrate the effect of collision-oriented guidance.
Without collision guidance, E2E-CDiff tends to maintain a natural and safe interaction with the ego vehicle.
After adding collision-oriented guidance, E2E-CDiff w/ coll still preserves plausible driving behavior at most time steps, but gradually steers the CBV toward the ego vehicle when the two vehicles enter a close interaction region.
This explains the higher CPK observed in Fig.~\ref{cpk}; collision-oriented guidance injects accident-prone interactions into otherwise natural behaviors, enabling E2E-CDiff w/ coll to construct safety-critical scenarios for vulnerability testing.



\begin{table*}[t]
\Large
\caption{The safety-critical evaluation of ADs under different CBV planners.}
\centering
{\resizebox{0.95\linewidth}{!}{
\begin{tabular}{c|cc|cc|cc|cc}
\toprule
\multirow{2}{*}{Method}&\multicolumn{2}{c}{PDM-Lite}&\multicolumn{2}{c}{PlanT}&\multicolumn{2}{c}{UniAD}&\multicolumn{2}{c}{VAD}\\
\cmidrule{2-3} \cmidrule{4-5}\cmidrule{6-7}\cmidrule{8-9}
&DS&RC&DS&RC&DS&RC&DS&RC\\
\midrule
PPO&89.77$\pm$5.18&92.75$\pm$3.45&45.43$\pm$2.10&65.91$\pm$0.72&71.22$\pm$4.61&\textbf{88.63$\pm$1.10}&69.42$\pm$4.43&82.33$\pm$1.61\\
Pluto&74.32$\pm$6.33&79.74$\pm$3.51&42.18$\pm$2.75&55.87$\pm$4.07&75.02$\pm$2.79&83.41$\pm$2.69&72.61$\pm$4.11&81.57$\pm$1.47\\
RIFT&93.23$\pm$2.19&94.42$\pm$1.61&46.76$\pm$2.01&62.83$\pm$4.08&\textbf{77.37$\pm$2.01}&87.84$\pm$4.08&75.67$\pm$1.80&83.42$\pm$0.39\\
\midrule
E2E-CDiff &\textbf{96.49$\pm$0.10}&\textbf{96.49$\pm$0.10} &\textbf{51.52$\pm$6.12}&\textbf{66.84$\pm$1.62}&69.59$\pm$1.95&85.99$\pm$4.71&\textbf{76.05$\pm$1.59}&\textbf{84.28$\pm$0.21}\\
\bottomrule
\end{tabular}
}}
\label{tablescenario1}
\end{table*}

\begin{table}[t]
\Large
\caption{Comparison between two-stage trajectory-then-PID and end-to-end control.}
\centering
{\resizebox{0.95\linewidth}{!}{
\begin{tabular}{c|cc|cc}
\toprule
\multirow{2}{*}{Method}&\multicolumn{2}{c}{Driving performance}&\multicolumn{2}{c}{Controllability}\\
\cmidrule{2-5}
&DS&RC&2D-TTC$\uparrow$ &ORR$\downarrow$\\
\midrule
Two-stage trajectory-then-PID&95.08$\pm$2.48&96.41$\pm$0.21&2.68$\pm$1.27&6.22$\pm$0.03\\
End-to-end control&96.49$\pm$0.10&96.49$\pm$0.10&2.94$\pm$1.24& 2.17$\pm$1.47  \\
\bottomrule
\end{tabular}
}}
\label{tables2stage}
\end{table}

\begin{figure}[t]
  \centering
\includegraphics[width=0.99\linewidth,scale=1]{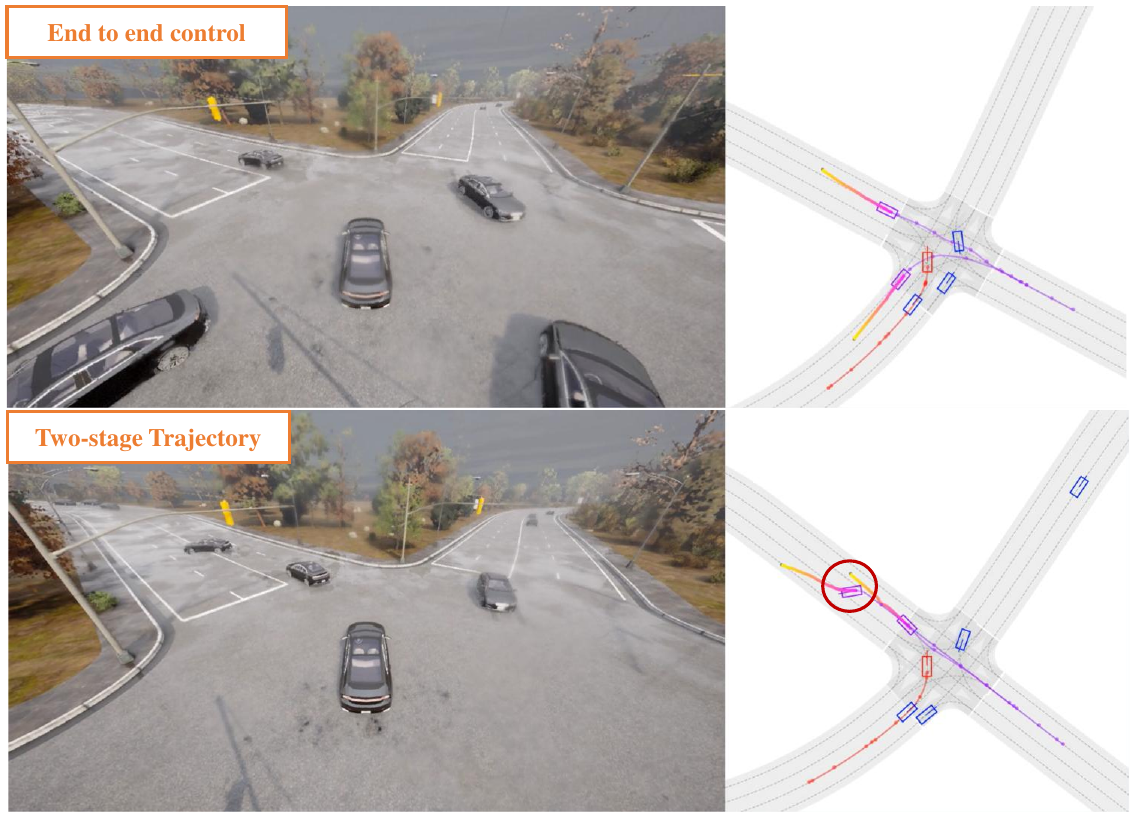}
  \caption{End-to-end control vs. two-stage trajectory-then-PID.}
  \label{e2e}
\end{figure}

\subsection{Ablation Study}
To address RQ3, we conduct ablation studies from two perspectives. First, we experimentally validate the effectiveness of end-to-end conditional diffusion in generating low-level control signals by comparing it with a two-stage trajectory-then-PID pipeline. Subsequently, we conduct ablation experiments to verify the controllability and realism of generated scenarios under different guidance conditions.

\subsubsection{End-to-end Control vs. Two-stage Trajectory-then-PID}
We evaluate the effectiveness of directly generating low-level control signals with the end-to-end conditional diffusion model by comparing it with a two-stage trajectory-then-PID pipeline.

Fig.~\ref{e2e} qualitatively compares the state-action rollouts generated by the two execution modes under the same scenario.
The two-stage pipeline first predicts a future state trajectory and then uses a PID controller to convert it into throttle, steering, and braking commands.
However, because low-level controls are computed after trajectory prediction and are not jointly modeled with the future states, the executed vehicle states may lag behind or deviate from the predicted trajectory, leading to a planning-control mismatch.
In contrast, end-to-end control jointly generates future states and executable actions within the same model, reducing the separation between motion prediction and control generation.
Table~\ref{tables2stage} further reports the quantitative comparison.
End-to-end control achieves higher DS and RC, improves 2D-TTC, and substantially reduces ORR compared with the two-stage trajectory-then-PID baseline.
These results demonstrate that directly modeling state-action trajectories improves closed-loop execution and controllability compared with the two-stage trajectory-then-PID baseline.

\subsubsection{Conditional Guidance Analysis}
Table~\ref{physical1} reports the ablation results of different guidance objectives, including collision-avoidance, drivable-area, and speed guidance.
The results show that these guidance terms play complementary roles in controlling CBV behavior.
Drivable-area guidance is important for lane compliance: variants without this guidance show much higher off-road rates, such as $7.05$ and $6.25$, while enabling drivable-area guidance reduces ORR to $3.50$ in the single-guidance setting.
Collision-avoidance guidance helps regulate interaction safety by preventing overly close ego--CBV encounters, while speed guidance constrains the generated motion to avoid aggressive or unrealistic velocity profiles.
When all guidance objectives are removed, the generated scenarios exhibit the lowest 2D-TTC and the highest ORR, indicating degraded controllability and rule compliance.
These results demonstrate that conditional guidance is essential for steering the diffusion model toward safe, drivable, and naturalistic CBV behaviors in closed-loop scenario generation.


\begin{table}[t]
\Large
\caption{Ablation study of conditional guidance objectives with PDM-Lite as the ego-vehicle planner.}
\centering
{\resizebox{\linewidth}{!}{
\begin{tabular}{ccc|ccccc}
\toprule
\multicolumn{3}{c}{Guidance Objectives}&\multicolumn{2}{c}{Controllability}&\multicolumn{3}{c}{Realism}\\
Collision&Drivable&Speed&2D-TTC$\uparrow$ &ORR$\downarrow$&S-SW$\uparrow$&A-SW$\uparrow$&S-WD$\downarrow$\\
\midrule
$\checkmark$&$\checkmark$&$\checkmark$&2.79$\pm$1.33&1.26$\pm$1.47&0.97$\pm$0.01&0.88$\pm$0.09&4.80$\pm$1.38\\
$\checkmark$&$\times$&$\times$&2.75$\pm$1.41&7.05$\pm$2.73&0.96$\pm$0.01&0.90$\pm$0.00&4.16$\pm$0.02\\
$\times$&$\checkmark$&$\times$& 2.83$\pm$1.20&3.50$\pm$1.93&0.88$\pm$0.03&0.77$\pm$0.03&5.61$\pm$0.38\\
$\times$&$\times$&$\checkmark$&2.79$\pm$1.25&6.25$\pm$2.11&0.89$\pm$0.11&0.83$\pm$0.13&5.11$\pm$1.50\\
$\times$&$\times$&$\times$&2.62$\pm$1.28&7.31$\pm$3.18&0.95$\pm$0.01&0.89$\pm$ 0.02&4.35$\pm$0.24\\
\bottomrule
\end{tabular}
}}
\label{physical1}
\end{table}
\begin{table}[t]
\Large
\caption{The performance comparison of ego-vehicle planners.}
\centering
{\resizebox{0.6\linewidth}{!}{
\begin{tabular}{c|cc}
\toprule
Method&DS&RC\\
\midrule
PDM-Lite&95.61$\pm$1.31&95.61$\pm$1.31\\
\midrule
PlanT&53.22$\pm$1.69&70.55$\pm$0.28\\
UniAD&76.14$\pm$1.29&88.96$\pm$0.63\\
VAD&78.76$\pm$2.14&83.03$\pm$0.09\\
\midrule
E2E-CDiff&79.96$\pm$1.63 &89.91$\pm$3.76\\
\bottomrule
\end{tabular}
}}
\label{tables6}
\end{table}
\begin{figure}[t]
  \centering
\includegraphics[width=0.9\linewidth,scale=1]{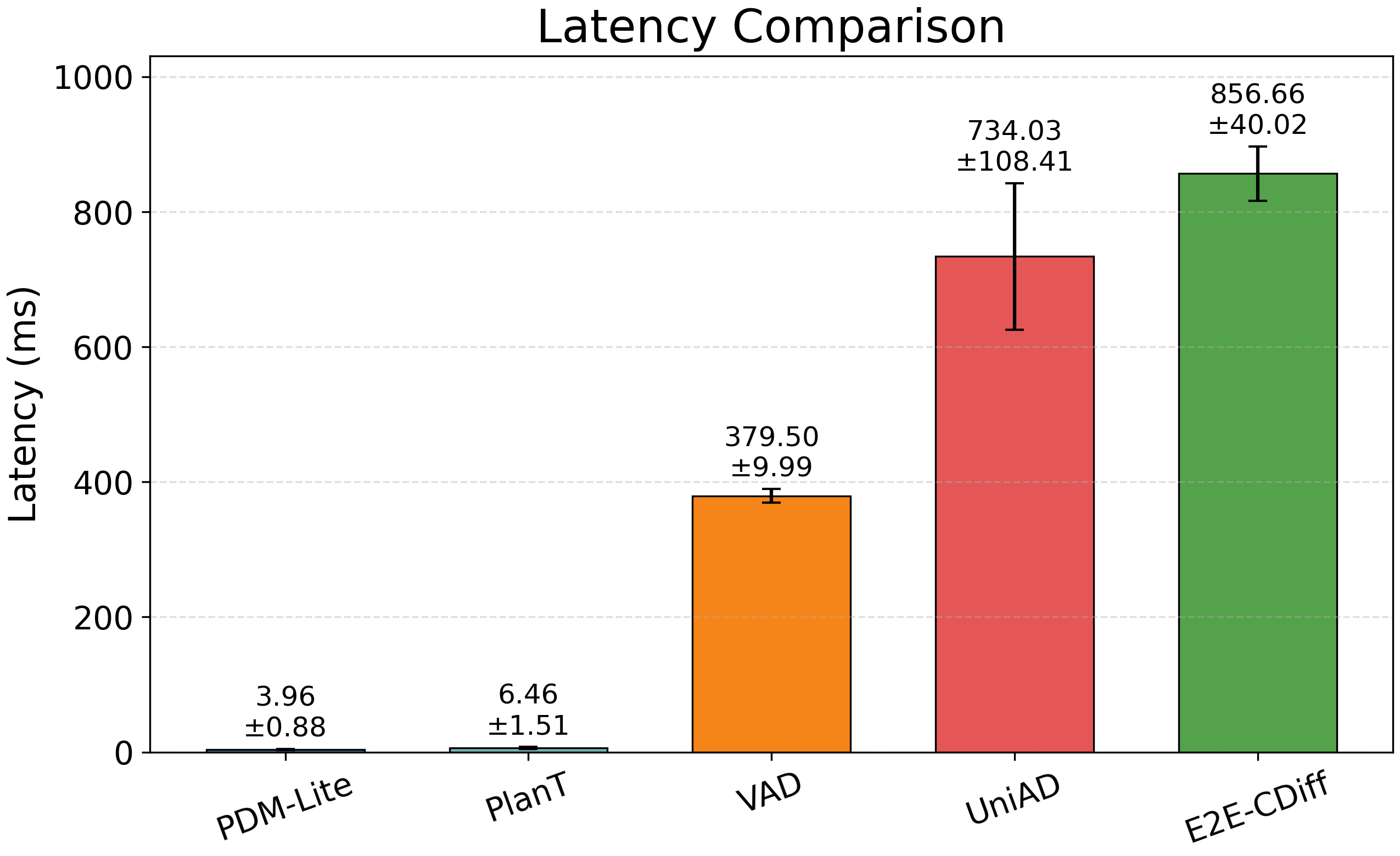}
  \caption{Latency comparison.}
  \label{latency}
\end{figure}
\subsection{E2E-CDiff as Ego-Vehicle Driving Planner}
To address \textbf{RQ4}, we further examine whether E2E-CDiff can be deployed as an ego-vehicle driving planner in addition to its primary use for controlling CBVs in traffic scenario generation.
Because E2E-CDiff generates an executable state-action rollout, it can directly output ego-vehicle control commands without relying on an additional trajectory-tracking controller.
We evaluate this capability from two perspectives: closed-loop driving performance and inference latency.
Here, latency denotes the end-to-end time from receiving the input observation and context to producing the control command for the current decision step.
\subsubsection{Ego-Vehicle Driving Performance}
We compare E2E-CDiff with representative ego-vehicle planners, including PDM-Lite, PlanT, UniAD, and VAD, using driving score (DS) and route completion (RC) as the main metrics.
As shown in Table~\ref{tables6}, E2E-CDiff achieves competitive closed-loop driving performance.
Although PDM-Lite obtains the best overall score due to its privileged rule-based design, E2E-CDiff performs favorably among learning-based planners and achieves higher DS and RC than several end-to-end baselines.
These results show that E2E-CDiff is not only effective for constructing controllable traffic scenarios through CBV control, but can also be directly used as an ego-vehicle driving planner.
Its competitive performance stems from its end-to-end state-action diffusion design, which generates future motion states together with executable low-level controls.
Moreover, conditional guidance provides additional structure during denoising, helping the generated actions remain controllable and consistent with closed-loop driving objectives.
\subsubsection{Latency Comparison}

Fig.\ref{latency} compares the average inference latency of different ego planners.
PDM-Lite and PlanT are highly efficient, requiring only 3.96 $\mathrm{ms}$ and 6.46 $\mathrm{ms}$, respectively.
VAD and UniAD require 379.50 $\mathrm{ms}$ and 734.03 $\mathrm{ms}$, while E2E-CDiff has the highest latency of 856.66 $\mathrm{ms}$.
This latency gap mainly comes from the iterative denoising process: instead of producing a control decision through a single forward inference, E2E-CDiff repeatedly refines a sampled state-action rollout.
Therefore, although E2E-CDiff achieves competitive driving performance, its current sampling cost remains a limitation for real-time ego-vehicle deployment, where delayed control responses can be unsafe in dynamic traffic interactions.
For closed-loop traffic scenario generation, however, this latency is less restrictive because the simulator can be synchronized with the planner at each decision step.
In this setting, the strengths of E2E-CDiff become more relevant: diffusion-based generation preserves realistic state-action behaviors, while conditional guidance provides controllability for constructing naturalistic or safety-critical scenarios.
Thus, E2E-CDiff is more suitable in its current form as a controllable and realistic scenario generation model, and improving diffusion sampling efficiency remains an important direction for future real-time ego-planner deployment.
\section{Conclusion}
This paper presented E2E-CDiff, an end-to-end conditional diffusion framework for controllable and realistic closed-loop traffic scenario generation.
Unlike conventional trajectory-then-controller pipelines, E2E-CDiff jointly models future vehicle states and executable low-level controls, reducing the planning-control mismatch during simulator execution.
Its differentiable guidance mechanism steers a shared generative model toward speed regulation, drivable-area compliance, collision-avoidance behavior, or collision-seeking behavior, enabling both naturalistic and safety-critical scenario generation without task-specific retraining.
Experiments on Bench2Drive demonstrate that E2E-CDiff achieves a favorable balance between controllability and realism and provides more reliable closed-loop execution than a two-stage trajectory-then-PID counterpart.
The collision-seeking variant further generates challenging interactions across multiple autonomous driving systems, highlighting the effectiveness of state-action diffusion as a flexible approach to scenario-based evaluation.

\bibliographystyle{IEEEtran}
\bibliography{tmm}

\end{document}